\documentclass[conference]{IEEEconf}
\newcommand{\radius}{\mathcal{R}}
\newcommand{\human}{\mathfrak{H}}
\newcommand{\net}{\mathtt{f}}
\newcommand{\real}{\mathbb{R}}
\newcommand{\inputx}{\vec{x}}
\newcommand{\inputy}{\vec{y}}

\newcommand{\labels}{\mathtt{C}}
\newcommand{\nothing}{\phi}
\newcommand{\setx}{\mathtt{X}}
\newcommand{\setxvalid}{\mathtt{X_c}}
\newcommand{\setxmis}{\mathtt{X_w}}
\newcommand{\mat}{W}
\newcommand{\bias}{\vec{b}}
\newcommand{\trans}{\mathtt{g}}

\newcommand{\algrobust}{\mathtt{isrobust}}

\newcommand{\algtest}{\mathtt{test}}

\newcommand{\algpvalue}{\mathtt{pvalue}}
\newcommand{\arxiv}[1]{}

\input{packages.tex}
\usepackage{algorithmic}
\usepackage{algorithm}

\renewcommand\thesection{\arabic{section}} 

\renewcommand\thesubsectiondis{\thesection.\arabic{subsection}}

\renewcommand\thesubsubsectiondis{\thesubsectiondis.\arabic{subsubsection}}

\begin{document}

\title{Input Validation for Neural Networks via  Local Robustness Verification}





\author{Jiangchao Liu$^{1}$,  Antoine Min\'{e}$^{2}$,
Hengbiao Yu$^{1}$, Ji Wang$^{1}$
\\
	\normalsize $^{1}$National University of Defense Technology, Changsha, China\\
	\normalsize $^{2}$Sorbonne Universit\'{e}, CNRS, LIP6, Paris, France\\
	\normalsize jiangchaoliu, hengbiaoyu, wj@nudt.edu.cn, Antoine.Mine@lip6.fr\\
}

\maketitle
\begin{abstract}
  Neural networks are known to be vulnerable to adversarial examples,
  which are inputs that are obtained by adding
  small, imperceptible perturbations to valid inputs,
  and that are designed to be misclassified.
  Local robustness verification can verify that a neural network
  is robust wrt. any perturbation to 
  a specific input
  within a certain distance. 
  This distance is called \emph{robustness radius}.
  We conducted empirical study on the robustness radii
  of different inputs. We  
  observed that the robustness
  radii of correctly classified inputs are much
  larger than that of natural misclassified (because of
  inaccuracy) inputs and 
  adversarial examples, especially those from strong
  adversarial attacks.
  Another observation is that the robustness radii
  of correctly classified inputs often follow a normal distribution.
  Based on these two observations, we propose to leverage local
  robustness verification techniques to 
  validate inputs for neural networks. Experiments show that our approach can
  protect neural networks from adversarial examples and improve their accuracy.
\end{abstract}
\IEEEoverridecommandlockouts
\vspace{1.5ex}

\maketitle

\section{Introduction}
\label{intro}

Despite the tremendous success~\cite{lecun2015deep}
of deep neural networks in recent years,
their applications in safety critical areas
are still concerning.

\emph{Can we trust neural networks?} This question arose when
people found it hard to explain or interpret neural networks~\cite{cast2016}
and drew more attention since the discovery of adversarial
examples~\cite{Szegedy13}.
An adversarial example is an input 
that is obtained by adding
a small, imperceptible perturbation to a valid input (i.e., correctly
classified input), and that is designed to be misclassified. 
Recent studies~\cite{ilyas2019} demonstrate
that adversarial examples are features
that widely exist in common datasets, thus
can hardly be avoided. This means neural networks inherently lack
robustness and are vulnerable to malicious attacks. 

Considerable amount~\cite{yuan2019} of works have been proposed to
improve the robustness of neural networks against adversarial examples.
One method is
\emph{adversarial training}~\cite{goodfellow2014}
which feeds adversarial examples to neural networks in the training stage.
Adversarial training works well on the types of adversarial examples considered
in the training dataset, but provides no guarantee on other types.
Some works~\cite{bradshaw2017} focus on designing robust architectures
of neural networks. However, similarly to adversarial training, these methods do
not guarantee robustness on all adversarial examples.

One promising solution
is \emph{formal verification}, which can prove that a network satisfies some
formally defined specifications.
To give a formal specification on robustness, we
first define a network as \(\net:\real^{m}\rightarrow \labels \), where
\(\real^{m}\) is a vector space of input (e.g., images) and 
\(\labels\) is a set of class labels. Then we define
\emph{robustness radius}~\cite{wang2017} of a network
 on an input \(\inputx \in \real^{m}\) as

\[
\radius(\net,\inputx) = \mathtt{max}\{\eta\mid \forall \inputy \in \real^{m},
\left\| \inputy - \inputx\right\|_{p} \leq \eta \Rightarrow \net(\inputx) =
\net(\inputy)\}
\]
where \(\left\| \cdot \right\|_{p}\) means \(L_{p}\) norm distance. Robustness
radius measures
the region in which a network is robust against perturbations. Another equivalent
definition is \emph{minimal distortion}~\cite{weng2018} (i.e., the minimal distance required
to craft an adversarial example). We prefer to use the term \emph{robustness
radius} since it is defined from a defensive perspective.
With robustness radius, we can define \emph{global robustness property} of
a network \(\net\) as

\[
\forall \inputx \in \real^{m}, \human(\inputx) = \net(\inputx) \Rightarrow
\radius(\net,\inputx) \geq \delta
\]
where  \(\delta\) is a user-provided threshold and \(\human:\real^{m}\rightarrow \labels\cup\{\nothing\}\) denotes the oracle on
the classification of \(\real^{m}\). Note that \(\human\) outputs \(\phi\) if  
an input in \(\real^{m}\) is not classified into any class in \(\labels\).
This property ensures that for any input that can be recognized by human
and correctly classified, the neural network is robust to any perturbation
to some extent (i.e., \(\delta\) in  \(L_{p}\) normal distance).
Unfortunately, the global robustness property following this definition can
hardly be verified because of the huge input space and  absence of the
oracle \(\human\). Some researchers tried to verify
global robustness property of a weaker definition~\cite{katz2017}
(without \(\human\)), but only succeeded on very small networks
(i.e., consisting of a few dozens of neurons). 

Given the difficulties in verifying global robustness properties,
many researchers turned to \emph{local robustness properties}, i.e.,
\(
\forall \inputx \in \setx\subseteq\real^{m}, \human(\inputx) = \net(\inputx) \Rightarrow
\radius(\net,\inputx) \geq \delta
\). Instead of the whole input space, local robustness property only considers
a set of inputs (denoted as \(X\) in the formula), for instance, the training dataset. Various
techniques~\cite{huang2017,gehr2018}
have been successfully applied in this kind of verification. However,
local robustness properties are currently only used to evaluate the
robustness of a given network or a defense technique, since
they do not provide guarantee for robustness on inputs
outside of the set \(\setx\).

\emph{Can we trust  neural networks on a specific runtime input?}
Although this question is a compromise to the sad fact that
the global robustness properties can hardly be guaranteed, it is still practically
useful if we can know whether a neural network gives the expected output on an
input at runtime.
\emph{Runtime verification}~\cite{desai2018} checks whether an output satisfies some
safety specifications at runtime and drops the output if not (traditional software is
used as backup).
This method, however, needs to know the constraints on
outputs, which is not the case in tasks like image classification.
\emph{Input reconstruction}~\cite{MengMagnet2017} tries to transform
adversarial examples to the inputs that can be correctly classified.
\emph{Adversarial detection}~\cite{lu2017}
rejects inputs that are suspected of being adversarial examples
based on the characteristics observed on known adversarial attacks.

Conventional local robustness verification needs an oracle on classification
and only measures the robustness radius of the correctly classified inputs.
However, the computation of robustness radius itself does not
need an oracle and can also be measured on natural misclassified inputs
(because of inaccuracy) and adversarial examples. 
It is known that adversarial examples themselves are often not robust to
small perturbations~\cite{luo2018,wang2018detect}.
We wonder whether
this implies that the robustness radii of adversarial examples should
be smaller than that of valid (i.e., correctly classified) inputs, and as a result,
robustness radii can be used as characteristics of inputs
for adversarial detection. To validate this idea,
we conducted empirical study on the robustness radii
of valid inputs, natural misclassified inputs and adversarial examples.
Our experiments are conducted on Feedforward
Neural Networks (FNN) and Convolutional Neural networks (CNN) with
three representative attacks, i.e., FGSM (fast, white-box)~\cite{goodfellow2014},
C\&W (strong, white-box)~\cite{carlini2017},
and  HOP (i.e., Hopskipjump, black-box)~\cite{chen2019}
on the datasets
MNIST~\cite{lecun1998} and
CIFAR10~\cite{cifar10}.
We have two observations.

The first observation is that the average robustness radius
of valid inputs (i.e., correctly classified inputs) is much larger than that of natural misclassified inputs,
and adversarial examples. To be formal,
given a neural network \(\net\), and a set of inputs  \(\setx\subseteq \real^{m}\)
(which may include adversarial examples),
let \(\setxvalid = \{\inputx\in \setx \mid \net(\inputx)= \human(\inputx) \}\)
be the set of valid inputs
and \(\setxmis = \{\inputx\in \setx \mid \human(\inputx) \in \labels \wedge \net(\inputx)\neq \human(\inputx)  \}\)
be the set of natural misclassified inputs and adversarial examples,
then we have

\begin{equation}
\label{observation}
\frac{\sum_{\inputx \in \setxvalid}\radius(\net,\inputx)}{\mid \setxvalid\mid}
\gg
\frac{\sum_{\inputx \in \setxmis}\radius(\net,\inputx)}{\mid \setxmis\mid}
\end{equation}
where \(\mid\cdot\mid\) denotes cardinality and \(\gg\) denotes
``much larger than''. Note that we only consider inputs that
can be classified into labels, which exclude randomly generated inputs mapping to
no label (i.e.,  mapped by \(\human\) to \(\nothing\)). We believe that
this assumption is practical.
Our experiments show that Equation~\ref{observation} holds
on adversarial examples from all attacks we have tried,
especially on those strong attacks which seek the smallest
perturbations.

Another observation is that the robustness radii of valid inputs
(i.e.,\(\{\radius(\net,\inputx)\mid \inputx \in \setxvalid\}\) ) often follow a normal
distribution. We say "often" because in our experiments, most (i.e., more than half)
cases passed normality test.

Based on these two observations, we propose to validate inputs for
neural networks via
robustness radius.
It can reject both
adversarial examples and natural misclassified  data which
often have small robustness radii. Thus it
not only protects neural networks from adversarial attacks,
but also improves their accuracies.
More importantly, this way does not need knowledge of the classification scenario
and is not specific to any attack. To be more specific, on a random CNN for
MNIST~\cite{lecun1998},
our method can reject 75\% misclassified natural inputs,
95\% and 100\% FGSM adversarial
examples with different parameters respectively, 100\% C\&W adversarial examples
and 100\% HOP adversarial examples, with only 3\% false alarm rate.

It is worth mentioning that the two observations are valid not only on
exact robustness radius computed by complete verification, but also on
under-approximated robustness radius computed by incomplete verification, which
is fast enough to be deployed  at runtime.

We make the following contributions:
\begin{itemize}
\item We conducted empirical study on the robustness radii of
  different categories of inputs on neural networks. We observed: (1)on FNNs and CNNs,
  the average robustness radius of the valid inputs
  is much larger than that of the misclassified clean inputs and adversarial examples;
  (2)on most FNNs and CNNs,
  the robustness radii of the valid inputs  follow a normal distribution;
\item Based on these two observations,
  we propose two new input validation methods based on local robustness
  verification (which currently is only used to evaluate the
  robustness of a given network, as opposed to validate inputs), which
  can protect neural networks from  adversarial examples, especially from
  strong attacks, and improve their accuracies on clean data.
\end{itemize}

\section{Background}

In this section, we give some background of the
current techniques for local robustness
verification of neural networks.

\subsection{Neural Networks}

In recent years, researchers have successfully verified local robustness properties
on various networks with different activation
functions~\cite{akintunde2019,singh2019}. However,
to highlight the main idea of  this paper,
we only consider Convolutional Neural networks (CNN) made of
fully-connected, convolutional and max-pooling layers, and ReLU
(i.e., \(ReLU(\inputx) = max(\inputx,0)\), where
\(max\) is applied point-wisely) as activation functions. A max-pooling
layer is equivalent to a composition of several fully-connected
layers~\cite{carlini2018}.
Thus we can safely consider only convolutional and fully connected layers
in our mathematical illustration.

Let \(\inputx_i\) be the  value vector of the neurons
in the \(i\)th layer, given
the input as \(\inputx_0 \in \real^{m}\).
Both convolutional and
fully connected layers can be denoted as a composition of
linear transformations and ReLU activations.
We use \(\trans_i\) to denote the transfer function
for  the \(i\)th layer,  then we have
\(\inputx_i = \trans_i(\inputx_{i-1}) =\)
\(ReLU(\mat_i\inputx_{i-1}+\bias_{i}) =\)
\(max(\mat_i\inputx_{i-1}+\bias_{i},0)\),
where \(\mat_i\) is the corresponding weights matrix and \(\bias_i\) is the
vector of bias.
Let \(\trans= \trans_n\circ \trans_{n-1}\circ\ldots\circ\trans_{1}\),
a neural network is defined as \(\net::=argmax\circ\trans\).

\subsection{Local Robustness Verification}

Recall the definition of local robustness property that

\[
\forall \inputx \in \setx\subseteq\real^{m}, \human(\inputx) = \net(\inputx) \Rightarrow
\radius(\net,\inputx) \geq \delta
\]

It ensures that, the neural network is immune to adversarial examples
on a set of inputs within \(\delta\) in \(L_{p}\) norm  distance.
To prove it, we only need to  prove that,
for given \(\inputx \in \setx \) and \(\delta\),

\begin{equation}
  \label{eq1}
\forall \inputy \in \real^{m}, \left\|\inputy - \inputx\right\|_{p} \leq \delta
\Rightarrow \net(\inputx) =
\net(\inputy)
\end{equation}


In this paper, we only consider the case \(p = \infty\).
Current verifiers for this property can be categorized as
\emph{complete} and \(incomplete\). Complete verifiers can give an
exact answer on whether the property is satisfied. However,
incomplete verifiers only provide conservative answers. To
be specific, since such verifiers
over-approximate the behaviors of neural networks, when their
over-approximation satisfies the property,
they return "true" (i.e., the neural networks must satisfy the
property), while if the over-approximation does not satisfy
the property, they can only return "unknown" rather than
"false".

\begin{figure}[!tbp]
\vskip 0.2in
\begin{center}
  \subfigure[Valid data]{\includegraphics[width=0.4\columnwidth]{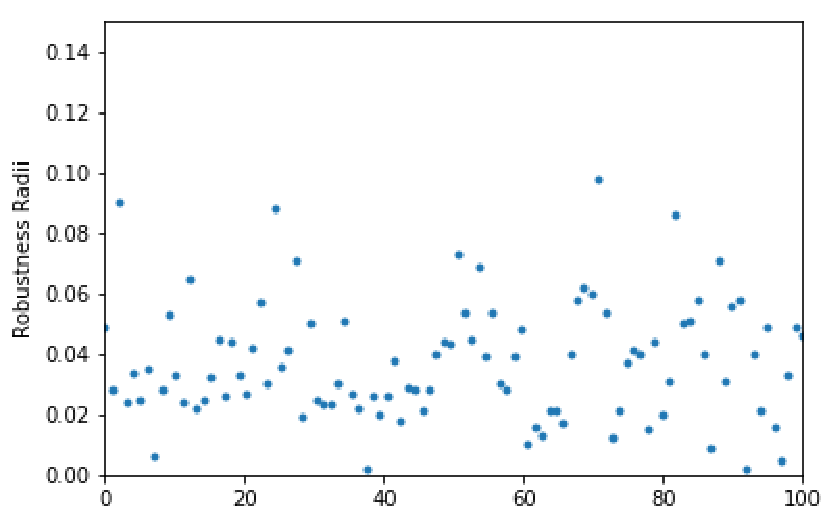}}
     \subfigure[Misclassified data]{\includegraphics[width=0.4\columnwidth]{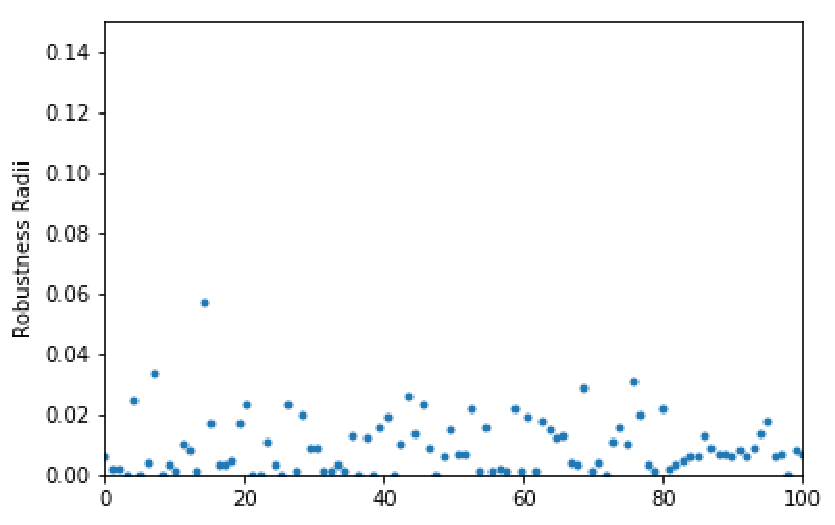}}
  \subfigure[Adversarial examples from FGSM with \(\epsilon = 0.1\)]{\includegraphics[width=0.4\columnwidth]{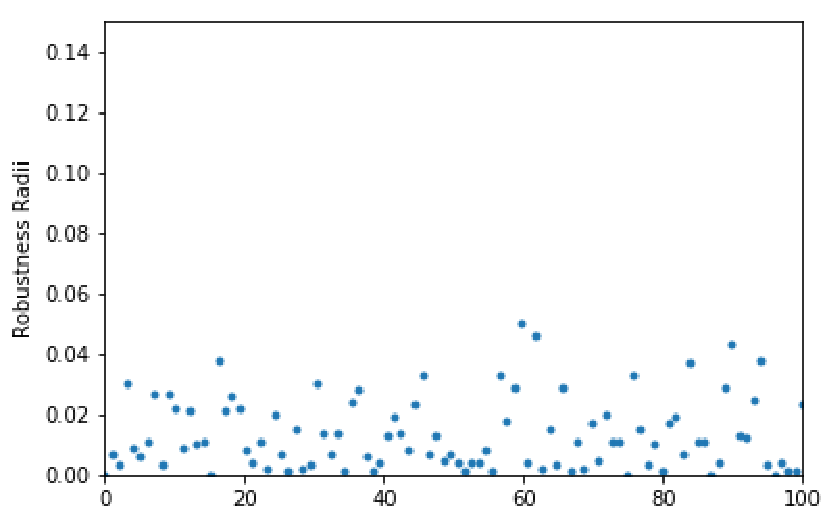}}
 \subfigure[Adversarial examples from FGSM with \(\epsilon = 0.05\)]{\includegraphics[width=0.4\columnwidth]{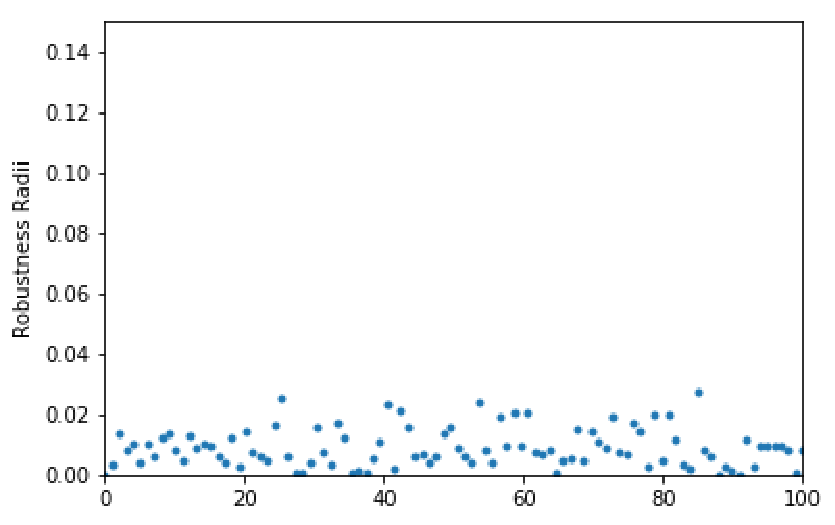}}
  \subfigure[Adversarial examples from C\&W]{\includegraphics[width=0.4\columnwidth]{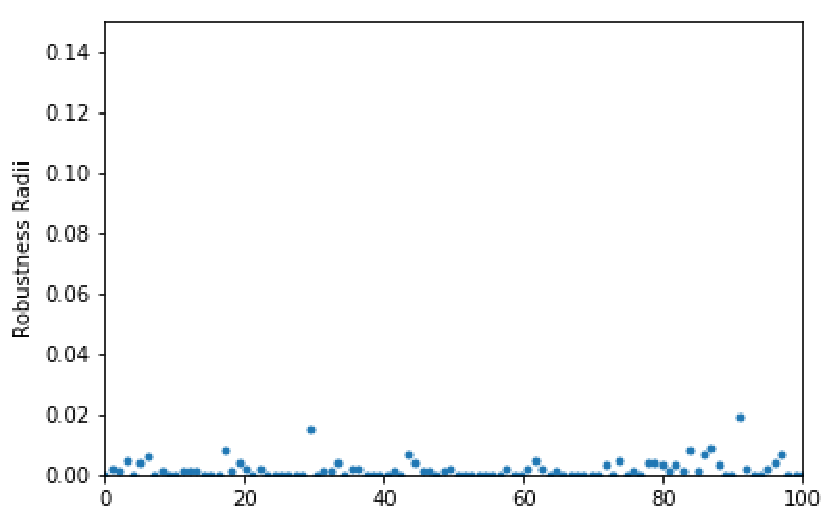}}
  \subfigure[Adversarial examples from HOP]{\includegraphics[width=0.4\columnwidth]{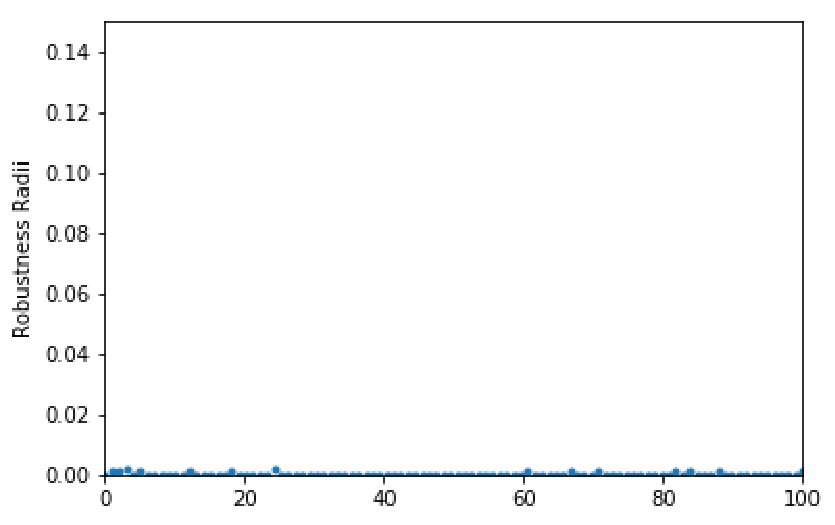}}
\caption{Robustness Radius from Complete Verification}
\label{Mnist-complete}
\end{center}
\vskip -0.2in
\end{figure}

\section{Observation on Robustness Radii of Inputs from Different Categories}

In this section, we conduct empirical study on
robustness radii of
of valid (i.e., correctly classified) data, natural misclassified
data and adversarial examples.

\emph{Experimental Setup}.
We take MNIST~\cite{lecun1998} and
CIFAR10~\cite{cifar10} as our input datasets and
use IBM's Adversarial Robustness Toolbox~\cite{art2018} to generate
FGSM,  C\&W, and  HOP adversarial examples with default parameters,
except for FGSM where we set \(\epsilon\) (i.e., a parameter~\cite{goodfellow2014}) as
0.1 (by default) and 0.05 (which is stronger) respectively.

We use ERAN~\cite{singh2018fast}
as the verifier which supports
both complete and incomplete robustness verifications. ERAN
does not  compute  robustness radius directly, but
can judge whether the robustness radius is larger than
a given value \(\delta\) (i.e., the network \(\net\) is robust on all inputs that
are within
\(\delta\) in  \(L_{\infty}\) norm distance with \(\inputx\), as Equation~\ref{eq1}).
We denote it as \(\algrobust(\net,\inputx,\delta)\).
Note that, ERAN supports two versions of \(\algrobust\): the complete one and
the incomplete one.
Applying binary search on the complete 
\(\algrobust\), we can find a value close enough
to the robustness radius.
In the incomplete version, \(\algrobust\) searches
for the boundary where ERAN  returns "true" and
"unknown" (rather than "true" and "false"), which
is  an underapproximation of the robustness radius.
This algorithm is described in Algorithm~\ref{alg:binary}.
It accepts four inputs: a network, an input sample, a big value  \(up\)
(larger than the maximal possible robustness radius) and a tolerance \(e\).
The output is a value,  the maximal
distance between  which and the (underapproximated) robustness radius is less than the tolerance.
In the following, we will  call the computed value with
complete verification the (asymptotically) \emph{exact robustness radius}, and
that with
incomplete verification the \emph{approximate robustness radius}.
All experiments are conducted on an Ubuntu 18.04 running on a desktop with an
Intel i9-9900K CPU, 32GB Memory.

\begin{algorithm}[tb]
   \caption{Computation of robustness radius}
   \label{alg:binary}
\begin{algorithmic}
  \STATE {\bfseries Input:} network $\net$, input \(\inputx\), big value \(up\), tolerance \(e\)
  \STATE Initialize \(low = 0\)
   \REPEAT
   \STATE \(mid = (up + low) / 2\)
   \IF{\(\algrobust(\net,\inputx,mid)\)}
   \STATE \(low = mid\)
   \ELSE
   \STATE \(up = mid\)
   \ENDIF
   \UNTIL{\(up - low < e\)}
   \STATE {\bfseries Output:} low
\end{algorithmic}
\end{algorithm}

\subsection{Observation on Exact Robustness Radius}
\label{obcomplete}

ERAN combines abstract interpretation,
linear programming and MILP to
completely verify a network.
To make the verification terminate in a reasonable time,
we trained a small FNN (denoted as FNN-MNIST) on MNIST
(with \(95.82\%\) accuracy), which consists of 5 layers: the input layer,
three fully connected layers, each with 30 neurons and one output layer with 10 neurons.

We run ERAN with RefineZono~\cite{singh2018} domain, and
set \(up\)  as 0.256 and \(e\) as 0.001
in Algorithm~\ref{alg:binary}.
We computed the  robustness radii of the first 100 samples from each
of following six categories in the MNIST test dataset:

\begin{compactitem}
\item samples that can be correctly classified by the network
\item samples that are misclassifed by the network
\item adversarial examples from successful FGSM attacks with \(\epsilon = 0.1\)
\item adversarial examples from successful FGSM attacks with \(\epsilon = 0.05\)
\item adversarial examples from successful C\&W attacks 
\item adversarial examples from successful HOP attacks 
\end{compactitem}

The  results are shown in Figure~\ref{Mnist-complete}.
Each point in the figures represents the robustness radius
of an input from the six categories.
We can see that the average robustness radius
of valid data are much larger than the other categories
of data, that is misclassfied inputs and adversarial examples.
To be specific, the robustness radii of
most valid inputs are larger than 0.02, while the robustness radii
of most samples
from other categories are smaller than 0.02.
One interesting phenomenon is that
the robustness radius of  adversarial examples from FGSM  attack with
\(\epsilon = 0.1\) are much larger than those with \(\epsilon = 0.05\).
This implies that the "holes" for misclassification is smaller near
the valid inputs. This phenomenon is more significant on 
the "strong" white-box C\&W attack and
black-box HOP attack, 
the robustness radii of  
the adversarial examples from which  are very close to 0.
Though, this may due to the fact that
these attacks try to find an adversarial example which is
as close as possible to the original input, thus
the result adversarial example is also close to the boundaries of classification.

\begin{figure}[ht]
\vskip 0.2in
\begin{center}
 \includegraphics[width=1\columnwidth]{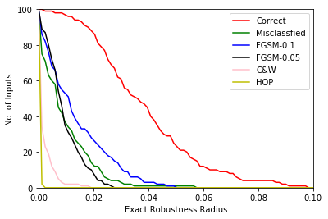}
 \caption{ The numbers of inputs which have a larger exact robustness radius  on FNN-MNIST than a given value }
 \label{ob1fnnc}
\end{center}
\vskip -0.2in
\end{figure}

Our experiments suggest that
we can use robustness
radius to evaluate to what extent we should trust the output of a neural network
on a given input. By setting
a threshold to reject any input the robustness radius of which is lower, 
we can protect the neural network from adversarial examples and improve
its accuracy.
Figure~\ref{ob1fnnc} shows the number of inputs, 
the exact robustness radii of which are above a given value (i.e., the x-axis).
If we see the x-axis as threholds, then, the values on
the y-axis indicate the ratios of inputs from each categories can be accepted
at run-time. This figure also indicates that we have to suffer from false alarms
if we would like many misclassified inputs and adversarial examples to be rejected.
Actually, if we set the threshold as 0.008, then
68\% misclassfied inputs,  45\% FGSM (\(\epsilon=0.1\)) adversarial examples,
51 FGSM (\(\epsilon=0.05\)) adversarial examples, 97\% C\&W adversarial examples
and 100\% HOP adversarial examples would be rejected, with a false alarm rate at 4\%.

However, complete verification is time-consuming. In our experiments,
each call to function \(\algrobust\) takes 11s on average, even though our network
contains only 100 neurons. It seems that complete verification can
hardly be deployed at runtime, especially considering that
the running time of complete verification increases exponentially
with the number of neurons.

\begin{figure}[ht]
\vskip 0.2in
\begin{center}
  \includegraphics[width=1\columnwidth]{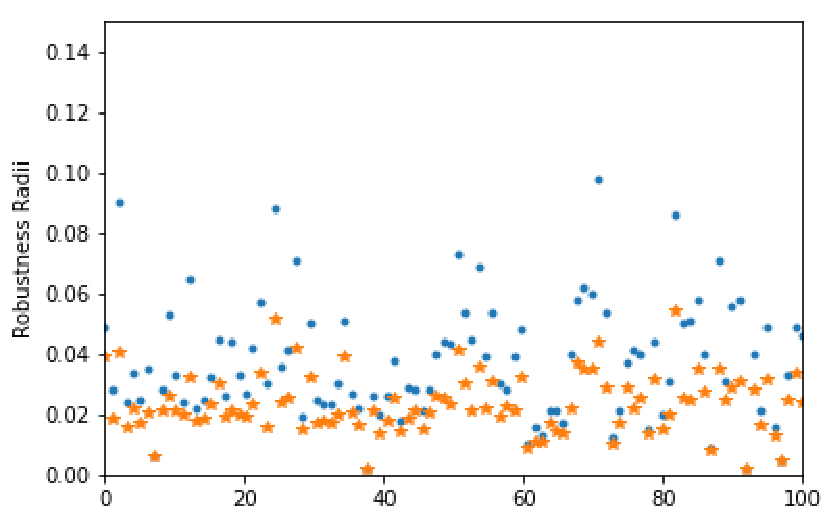}
\caption{The exact/approximate robustness radii of the first 100 valid inputs}
\label{Mnist-compare}
\end{center}
\vskip -0.2in
\end{figure}



\subsection{Observation on Approximate Robustness Radius}
\label{obincomplete}

Incomplete verification usually runs much
faster than complete verification and has the potential to be deployed
at runtime. However, Algorithm~\ref{alg:binary} with
incomplete verification can only give an approximate robustness
radius. We would like to know  (1) whether approximate robustness
radius from incomplete verification is close to the exact robustness radius;
(2) whether
the approximate robustness radii of valid inputs are significantly larger than
that of the misclassified inputs and adversarial examples.
Actually the second question is more important since
it decides whether we can use incomplete verification to validate inputs.


\emph{Observation on FNN}. We utilize ERAN with DeepZono domain~\cite{singh2018fast}
(which is incomplete)
to compute the approximate robustness radii of the same inputs
as Section~\ref{obcomplete} on the network FNN-MNIST.

The exact (resp. approximate)
robustness radii of the first 100 valid inputs in the form of
blue dot "." (resp. brown star "*") are shown in Figure~\ref{Mnist-compare}.
We can see that
the values of approximate and exact robustness radii of the same inputs are
very close. In fact, the approximate robustness radii (except those equal to 0)
of all inputs are between 44\% and 100\% of their exact robustness radii.

\begin{figure}[ht]
\vskip 0.2in
\begin{center}
  \includegraphics[width=1\columnwidth]{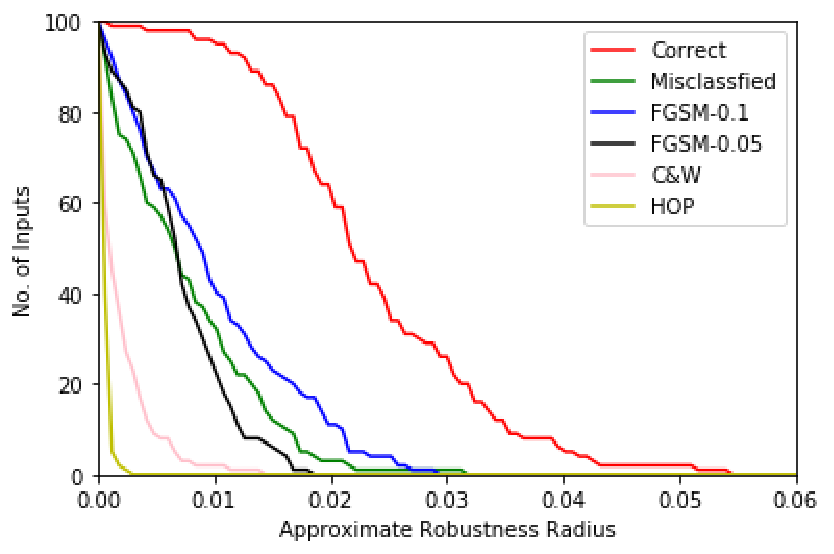}
  \caption{The numbers of inputs which have a larger approximate robustness radius on FNN-MNIST than a given value}
  \label{ob1fnninc}
\label{ob1}
\end{center}
\vskip -0.2in
\end{figure}

Figure~\ref{ob1fnninc}
shows the number of inputs, 
the approximate robustness radii of which are above a given value (i.e., the x-axis).
We can see that, similar to exact robustness radii,
the approximate robustness radii of  valid inputs are significantly
larger than that of misclassified inputs and adversarial examples, especially
those from "strong" attacks. Different from exact robustness radii which
measure the distance of the inputs with the boundaries of classification,
approximate robustness radii
reflect the accumulative gradient information of the neural network around
the input. Surprisingly, approximate robustness radii follows a similar distribution with
exact robustness radii, that is, they are smaller when the adversarial
examples are closer to the original inputs. This implies that the
misclassfied inputs and adversarial examples near the valid inputs
are more likely not to be robust.

Figure~\ref{ob1fnninc} also shows that, we can utilize
approximate robustness radius to protect neural networks from adversarial
examples and improve its precision.
Actually,  by  setting a threshold
as 0.012, 78\%
misclassified inputs,
67\%  adversarial examples from FGSM(\(\epsilon=0.1\)) attacks,
89\% adversarial examples from FGSM(\(\epsilon=0.05\)) attacks,
99\% adversarial examples from C\&W attacks,
and 100\% adversarial examples from HOP attacks can be rejected,
with only 7\% false alarm rates.
If we would like no valid input to be mistakenly rejected, by setting
the threshold as 0.001, we can still reject
14\% misclassified inputs,
7\%  adversarial examples from FGSM(\(\epsilon=0.1\)) attacks,
10\% adversarial examples from FGSM(\(\epsilon=0.05\)) attacks,
53\% adversarial examples from C\&W attacks,
and 90\% adversarial examples from HOP attacks,
with 0 false alarm.

Moreover, each call to \(\algrobust\) in incomplete verification costs less
than 1s on the given network, and has polynomial time complexity wrt. the number
of neurons, which means it has potential to be deployed at runtime.

\emph{Observation on CNN}.

We have also conducted experiments on Convolutional Neural Networks (CNN). They
are significantly larger than the network FNN-MNIST, such that complete
verification methods can hardly compute robustness radius in a reasonable time.
Thus we only tried incomplete verification.
Our experiments on CNN are conducted on two datasets: MNIST~\cite{lecun1998} and CIFAR10~\cite{cifar10}.

We trained a CNN (denoted as CNN-MNIST) on MNIST of 7 layers:
the input layer, a convolutional layer with 6 filters of size \(3\times 3\),
a max-pooling layer of \((2,2)\), a convolutional layer with 16 filters of size \(3\times 3\),
a max-pooling layer of \((2,2)\), a fully connected layer of 128 neurons and
an output layer with 10 labels. The accuracy is \(98.62\%\).

\begin{figure}[ht]
\vskip 0.2in
\begin{center}
\includegraphics[width=1\columnwidth]{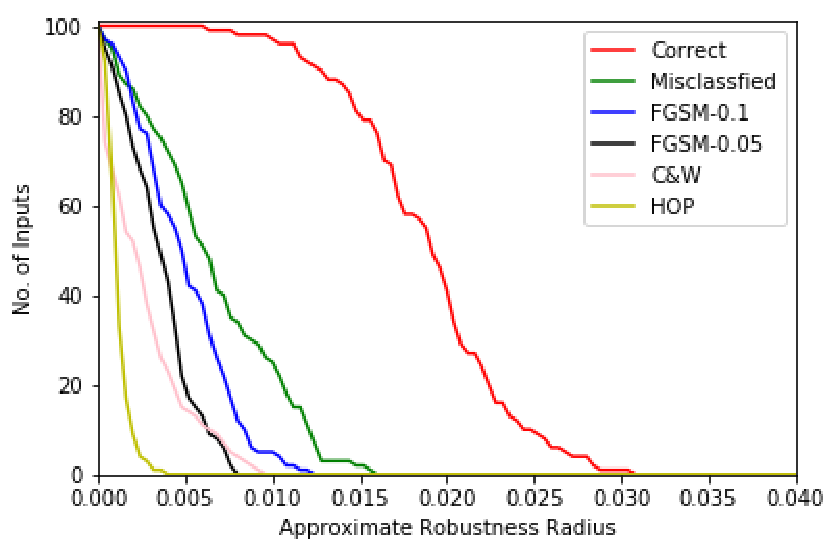}  
  \caption{The numbers of inputs which have a larger approximate robustness radius on CNN-MNIST than a given value}
  \label{ob1cnnmnist}
\end{center}
\vskip -0.2in
\end{figure}

As in Section~\ref{obcomplete}, we utilize ERAN with DeepZono
domain~\cite{singh2018fast}
to compute the approximate robustness radii of the first 100
inputs from each of the six categories.
Figure~\ref{ob1cnnmnist}
shows the number of inputs, 
the approximate robustness radii of which are above a given value.
We can see that the computed approximate robustness
radii of all inputs  are much smaller (i.e., \(\leq 0.04\)) than those
computed on the small network FNN-MNIST. We do not
know whether the approximate robustness radii are close to the
exact robustness radii, which we cannot get even after several days of
computation. 
However, most importantly, the characteristics of approximate robustness radii
of inputs of different categories are the same as exact robustness radii.
That is the approximate robustness radii of valid inputs (i.e., the red line) are much larger
than that of other inputs. In fact, if we set the threshold as 0.01, we can reject
\(75\%\) misclassified clean data, \(95\%\) FGSM adversarial examples where
\(\epsilon = 0.1\), \(100\%\) FGSM adversarial examples where
\(\epsilon = 0.05\),  \(100\%\) C\(\&\)W adversarial examples,
and \(100\%\) HOP adversarial examples, and only \(3\%\) valid
inputs.

\begin{figure}[ht]
\vskip 0.2in
\begin{center}
\includegraphics[width=1\columnwidth]{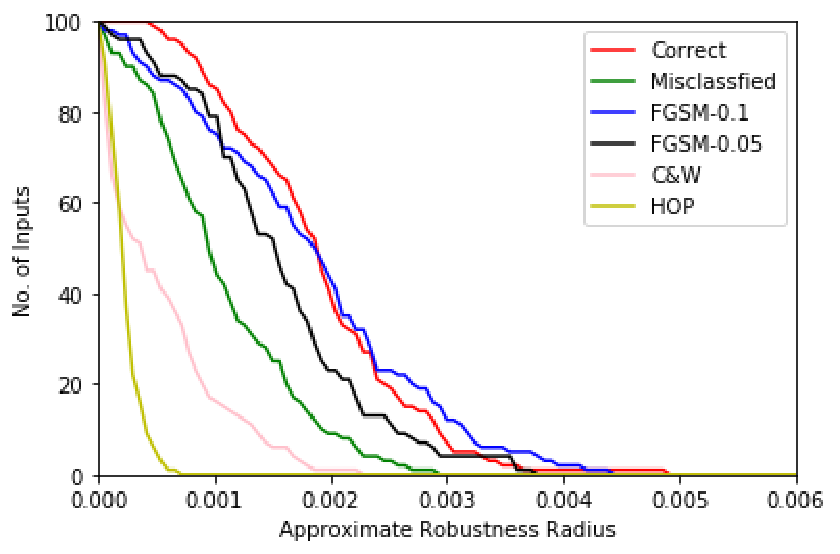}
  \caption{The numbers of inputs which have a larger approximate robustness radius on CNN-CIFAR than a given value}
  \label{ob1cnncifar}
\end{center}
\vskip -0.2in
\end{figure}


We trained a LeNet-5~\cite{lecun1998} CNN (denoted as CNN-CIFAR in this paper)
on CIFAR10  of 8 layers:
the input layer,
a convolutional layer with 6 filters of size \(5\times 5\),
a max-pooling layer of \((2,2)\), a convolutional layer with 16 filters of size
\(5\times 5\),
a max-pooling layer of \((2,2)\), two fully connected layers of 120 and 84 neurons
respectively,
an output layer with 10 labels.
The accuracy is \(73.66\%\).

Figure~\ref{ob1cnncifar} shows the results of the first 100
inputs of each category in CIFAR10 test database.
Even though the approximate robustness radii of the valid inputs
are significantly larger than those of misclassified
inputs and adversarial examples from C\&W and HOP attacks, but are
almost indistinguishable from FGSM attacks. We believe that the reason
is the accuracy of the network is too low such that it leaves big ``holes''
for adversarial examples in the input space.
There are CNNs~\cite{hu2018,xie2017} that have high accuracies on CIFAR10.
But  these networks usually adopt layers other than fully-connected,
convolutional and max-pooling layers and are out of the scope of this paper
and beyond the ability of current local robustness
verifiers~\cite{katz2017,wang2018,,singh2018fast}.
Thus we cannot testify our assumption.
However, our method is valuable at least
on small networks, which can also be deployed in real life.


To confirm our observations, we trained more FNNs and CNNs of various structures
and conducted the same measure. Table~\ref{p-table} shows the results. In the
table, on each network, we show its training dataset (column Dataset),
network structure (column Network), where (\(3\times30,10_{1}\)) describes
three fully connected layers of 30 neurons with an output layer of 10 neurons (i.e., FNN-MNIST), and
the subscript 1 is just to distinguish different networks with the same structure.
The network (\(6^{(3,3)},16^{(3,3)},128,10_{1}\)) describes
a CNN with a convolutional layer of 6 filter of 3\(\times\) 3,
a convolutional layer of 16 filters of 3\(\times\) 3,
a fully connected layer of 128 neurons and an output layer of 10 neurons (i.e., CNN-MNIST).
The network (\(6^{(5,5)},16^{(5,5)},120,84,10\)) describes CNN-CIFAR.
 For each network structure, we adopted different activation functions (column Activation).
 The table also shows the
accuracy on the test dataset (column Acc.) and the average approximate
robustness radii of the first 10 inputs 
in the test dataset of six categories:
correctly classified inputs (column Valid), misclassified inputs (column Mis.), adversarial
examples from FGSM attacks with \(\epsilon = 0.1\), FGSM attacks with \(\epsilon=0.05\),
C\&W  attacks and HOP attacks.
We chose 10 because we believe that 10 is enough to
compare the average values, and generating adversarial attacks e.g., HOP can be very time-consuming.
The average running time of each call to \(\algrobust\) is also
recorded in column Time(s).
The column P-value will be explained later (see Section~\ref{distribution}).
From the table, we can see that our observation is valid on all trained networks.
That is, the approximate robustness radii of valid inputs are much more than
misclassfied inputs and adversarial examples, especially those from
strong attacks. The running time for each call
to  \(\algrobust\)  is also short enough (mostly below 1s) to be deployed
at runtime.

Our experiments can be easily reproduced since we only use open source
tools with a little modification
(e.g., Algorithm~\ref{alg:binary}).
The modified code and all trained networks in this paper
have been uploaded as complementary materials.
Even though we believe
that people can easily reproduce our experiments with their own trained networks.

\begin{table*}[t]
  \caption{The average robustness radius and p-value of different networks}
\label{p-table}
\vskip 0.15in
\begin{center}
\scriptsize
  \begin{sc}
    \begin{tabular}{ccccccccccccc}
    \hline
    \hline
Dataset & Network & Acivation& Acc.& Valid & Mis. & FGSM(\( 0.1\)) & FGSM(\(0.05\)) & C\& W &HOP & P-value  & Time(s)\\
MNIST & \(3\times30,10_{1}\)            &ReLU & 95.82 &0.0227&0.0056 & 0.0091 & 0.0066&0.0020&0.0009&  0.033 & 0.242\\
MNIST & \(3\times30,10_{2}\)           &ReLU & 96.49 &0.0183& 0.0069 & 0.0087 & 0.0065 &  0.0020  & 0.0009& 0.026  &   0.239\\
MNIST & \(3\times30,10_{3}\)           &ReLU& 96.55 &0.0194& 0.0069 &  0.0078& 0.0048 & 0.0027  & 0.0007 &  \(<\)0.001 & 0.241 \\
MNIST & \(3\times50,10\)               &ReLU& 96.80 & 0.0176& 0.0047  & 0.0090 & 0.0068& 0.0013  & 0.0009 &   0.140 &  0.143 \\
MNIST  & \(3\times100,10\)              &ReLU& 97.49 & 0.0162 & 0.0070&  0.0080  &0.0054 & 0.0023  & 0.0010   & 0.916    & 0.357 \\
MNIST & \(4\times100,10\)              &ReLU& 97.37 & 0.0143 & 0.0076&  0.0057 & 0.0051  & 0.0019 &0.0010& 0.091  &  0.441 \\
MNIST & \(5\times100,10\)              &ReLU & 97.61 & 0.0127 &0.0057 & 0.0047 & 0.0043 &  0.0013 &  0.0011 & 0.082 & 0.517 \\
MNIST & \(5\times200,10\)              &ReLU& 97.70 & 0.0101 & 0.0054 & 0.0038 & 0.0041 &  0.0027 &  0.0010& \(<\)0.001 & 1.746 \\
MNIST & \(3\times30,10\)               &Sigmoid& 95.26 &0.0145  &0.0071  &  0.0074  &0.0050 & 0.0020 & 0.0011  &    \(<\)0.001  &      0.251 \\
MNIST & \(3\times50,10\)               &Sigmoid& 96.15& 0.0130 & 0.0053 & 0.0071 &0.0050 &   0.0015 &   0.0010 &  0.003&  0.156 \\
MNIST & \(3\times100,10\)              &Sigmoid& 97.11 & 0.0115& 0.0057 &0.0047 & 0.0043 & 0.0017 & 0.0008  & 0.142  & 0.408  \\
MNIST & \(4\times100,10\)              &Sigmoid& 96.86 & 0.0101& 0.0057 &0.0041 & 0.0042&  0.0007 & 0.0008  & 0.126  &  0.551  \\
MNIST & \(5\times100,10\)              &Sigmoid& 96.30& 0.0102& 0.0023&  0.0050& 0.0024 &  0.0011 &   0.0008  & 0.492  & 0.670 \\
MNIST & \(5\times200,10\)              &Sigmoid& 96.90 & 0.0086& 0.0052 &  0.0042 & 0.0033  & 0.0018 &  0.0007 &  0.034 & 2.720  \\

MNIST & \(3\times30,10\)               &Tanh&    96.25 & 0.0081& 0.0046&  0.0030 & 0.0020 & 0.0016 & 0.0008& \(<\)0.001 &   0.255 \\
MNIST & \(3\times50,10\)               &Tanh&    96.98 & 0.0070& 0.0032 &  0.0034& 0.0034 & 0.0008 & 0.0007 & \(<\)0.001 &  0.156 \\
MNIST & \(3\times100,10\)              &Tanh&    97.42& 0.0059 & 0.0024 &  0.0034 & 0.0026& 0.0011 & 0.0007& 0.713   &  0.408 \\
MNIST & \(4\times100,10\)              &Tanh&    97.80& 0.0046&   0.0026   & 0.0017 &0.0021 &  0.0013 & 0.0008  &   0.307    & 0.550 \\
MNIST & \(5\times100,10\)              &Tanh&    97.62 &0.0035 &  0.0020 &  0.0024  & 0.0016& 0.0008  & 0.0007   & 0.022    & 0.681  \\
MNIST & \(5\times200,10\)              &Tanh&    97.52 & 0.0028&   0.0015   & 0.0015 &  0.0010  &   0.0007   & 0.0007   & 0.142  & 2.736 \\

MNIST & \(6^{(3,3)},16^{(3,3)},128,10_{1}\) &ReLU& 98.62& 0.0175&  0.0064   & 0.0055  & 0.0032  & 0.0021 &   0.0011  &    0.828 &  0.232\\
MNIST & \(6^{(3,3)},16^{(3,3)},128,10_{2}\)&ReLU & 98.29& 0.0188& 0.0061  &   0.0049 & 0.0040   &   0.0034    & 0.0019  &   0.492&  0.253 \\
MNIST & \(6^{(3,3)},16^{(3,3)},128,10_{3} \)&ReLU& 98.45 &0.0174&0.0065 & 0.0051 & 0.0029 &    0.0030   & 0.0019 &0.622   &  0.213 \\

MNIST & \(3^{(3,3)}, 8^{(3,3)},128,10\) &ReLU& 97.45    & 0.0239& 0.0092    &  0.0089&  0.0080 &  0.0025 &   0.0021 &   \(<\)0.001& 0.073\\
MNIST & \(12^{(3,3)},32^{(3,3)},128,10\) &ReLU & 98.99  & 0.0170&  0.0059 &   0.0051 & 0.0032 & 0.0027  & 0.0019 & 0.333  &   0.754 \\
MNIST & \(6^{(3,3)},16^{(3,3)},128,10 \)& Sigmoid &98.45& 0.0178   & 0.0045   & 0.0053& 0.0045 & 0.0017&  0.0018&0.773  & 0.344\\
MNIST & \(3^{(3,3)}, 8^{(3,3)},128,10\) &Sigmoid& 97.78 &0.0249  & 0.0081 & 0.0102 & 0.0060&  0.0037  & 0.0022  & 0.068  &  0.119\\
MNIST & \(12^{(3,3)},32^{(3,3)},128,10\) &Sigmoid& 98.85 & 0.0172 & 0.0061 & 0.0038 & 0.0059 & 0.0016   & 0.0019& 0.522  & 1.318\\
MNIST & \(6^{(3,3)},16^{(3,3)},128,10 \)& Tanh & 98.73 & 0.0039  & 0.0013 & 0.0022  &  0.0015 &  0.0006 & 0.0006  & 0.188 & 0.369 \\
MNIST & \(3^{(3,3)}, 8^{(3,3)},128,10\) & Tanh &98.56 &  0.0103  & 0.0036 &0.0042 &0.0024 &   0.0014    & 0.0012  &  0.314 &   0.124\\
MNIST & \(12^{(3,3)},32^{(3,3)},128,10\) &Tanh & 99.11 & 0.0037  & 0.0017&0.0024 & 0.0013 &    0.0004    & 0.0006 & 0.001 & 1.488\\

CIFAR10 & \(6^{(5,5)},16^{(5,5)},120,84,10\)& ReLU & 73.66 &  0.0017 &  0.0009  &     0.0017    &  0.0016  &  0.0003  &  0.0002 &0.005  & 1.680\\
CIFAR10 & \(6^{(5,5)},16^{(5,5)},120,84,10\)& Sigmoid & 65.30 &  0.0025  &  0.0014  &  0.0029  &    0.0022  &  0.0005   &   0.0004  & 0.839   & 2.418\\
CIFAR10 & \(6^{(5,5)},16^{(5,5)},120,84,10\)& Tanh & 70.68 &  0.0008 &  0.0005  &  0.0009  &   0.0007    &   0.0002 & 0.0002   &0.390   & 2.443\\
\hline\hline
\end{tabular}
\end{sc}
\end{center}
\vskip -0.1in
\end{table*}

\section{Input Validation with Observation I}

Based on our first observation, we  design an algorithm to validate
the inputs of a neural network at runtime to protect it from 
adversarial examples and improve its accuracy.

\begin{figure}[ht]
\vskip 0.2in
\begin{center}
  \subfigure[ROC curve on MNIST]{\includegraphics[width=0.49\columnwidth]{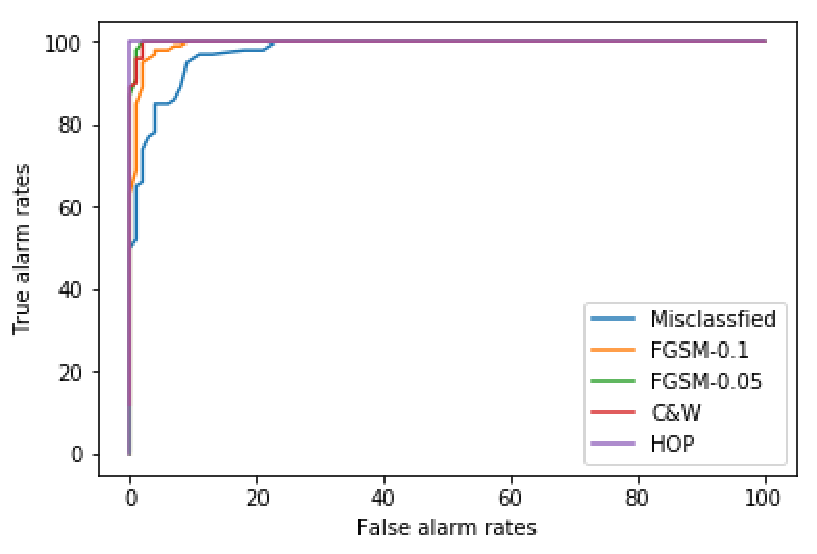}}
  \subfigure[ROC curve on CIFAR]{\includegraphics[width=0.49\columnwidth]{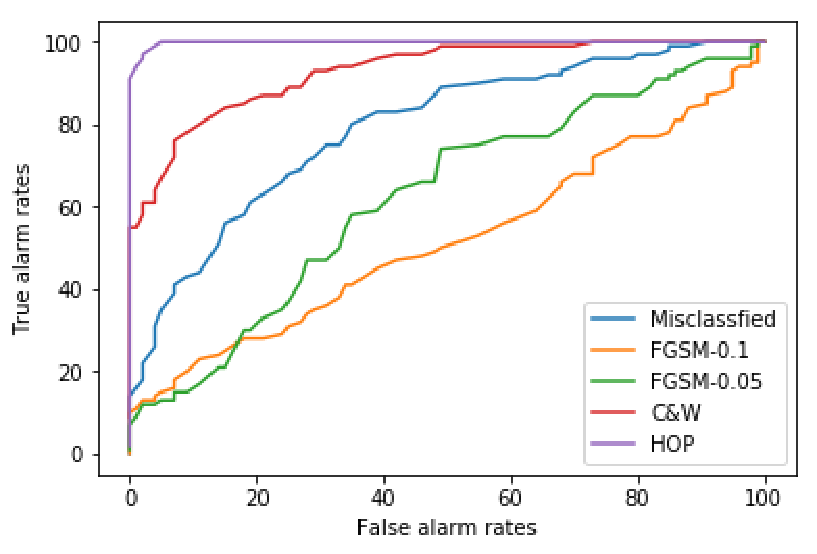}}
\caption{ROC curves on MNIST and CIFAR}
\label{roc-curve}
\end{center}
\vskip -0.2in
\end{figure}

A naive idea is to set a threshold and reject all inputs the approximate
robustness
radii of which are below it.
We call this method \emph{validation by threshold}.
However, it is non-trivial to choose the threshold.
One solution is to set the threshold according to the maximal false alarm
rate that can be tolerated, which
depends on the application.
ROC curve plots the true alarm rate against the false
alarm rate at various threshold settings. 
Figure~\ref{roc-curve} shows the ROC curves of the network
CNN-MNIST and CNN-CIFAR on the first 100 inputs from each category.
The result on MNIST
is good on all kinds of adversarial examples. However, on CIFAR10,
our method is not very helpful on FGSM attacks.
The reason, we believe, is that the accuracy of our CNN on CIFAR10 is
not high enough.

Until now, we have only studied the first 100 inputs in each category.
In Table~\ref{th-table}, 
we show the effect of different choices of thresholds
on the network CNN-MNIST  on the first
100 and random 100 inputs in each category.
To be more specific, we show  with different thresholds (column Th.),
the percentage of rejected valid inputs (column Vic.),
the percentage of rejected misclassified inputs (column W.), and the rejected adversarial examples from
FGSM attack with \(\epsilon = 0.1\) (column F (\(0.1\))),
FGSM attack with
\(\epsilon = 0.05\) (column F (\(0.05\))), C\&W attack
and HOP attack. The result of the first 100 inputs and random 100 inputs are
on the left and right sides of "/" respectively in each cell.
This table shows that the observation of the first 100 inputs of each category
are also valid in the whole test database.

\begin{table}[t]
  \caption{The rejection rates with different thresholds}
\label{th-table}
\vskip 0.15in
\begin{center}
  \scriptsize
  \begin{sc}
  \begin{tabular}{lcccccc}
Th.& Vic.& W&  F (\(0.1\)) &
F (\(0.05\))&
C\&W&
HOP\\
0.002 & 0/0 & 14/14 & 17/13 & 27/24 & 48/47 & 91/99\\
0.004 & 0/0 & 28/29 & 42/38 & 57/56 & 77/79 & 100/100\\
0.006 & 0/0 & 49/48 & 62/59 & 87/79 & 89/89 & 100/100\\
0.008 & 2/1 & 66/69 & 88/79 & 100/95 & 96/99 & 100/100\\
0.010 & 3/4 & 75/80 & 95/97 & 100/99 & 100/100 & 100/100\\
0.012 & 8/7 & 89/90 & 99/100 & 100/100 & 100/100 & 100/100\\
0.014 & 13/16 & 97/95  & 100/100 & 100/100 & 100/100 &100/100\\
0.016 & 24/28 & 100/100  & 100/100 & 100/100 & 100/100 &100/100\\
\end{tabular}
\end{sc}
\end{center}
\vskip -0.1in
\end{table}

The benefit of \emph{validation by threshold} is that once the
threshold is decided, we just need to  call \(\algrobust\) once to test
whether the approximate robustness radius of an input is above the threshold.

\section{Observation on The Distribution of  Approximate Robustness Radii of Valid Inputs}
\label{distribution}

One thing concerns us: if the attackers have the knowledge  of
our neural network
and our detection method, they can generate adversarial examples with large approximate
robustness radii on purpose (even though we believe that such adversarial examples can
be hardly found on a neural network with high accuracy). To avoid this, we 
study further whether the approximate robustness radii of valid inputs follow a certain
distribution. If they do, then the attackers not only need to generate adversarial
examples with large enough robustness radii, but also need to make sure that such 
robustness radii follow a certain distribution, which is much harder.

\begin{figure}[ht]
\vskip 0.2in
\begin{center}
  \subfigure{\includegraphics[width=0.48\columnwidth]{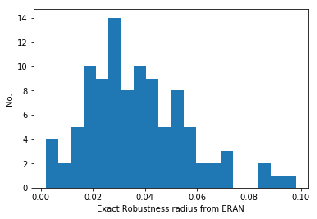}}
  \subfigure{\includegraphics[width=0.48\columnwidth]{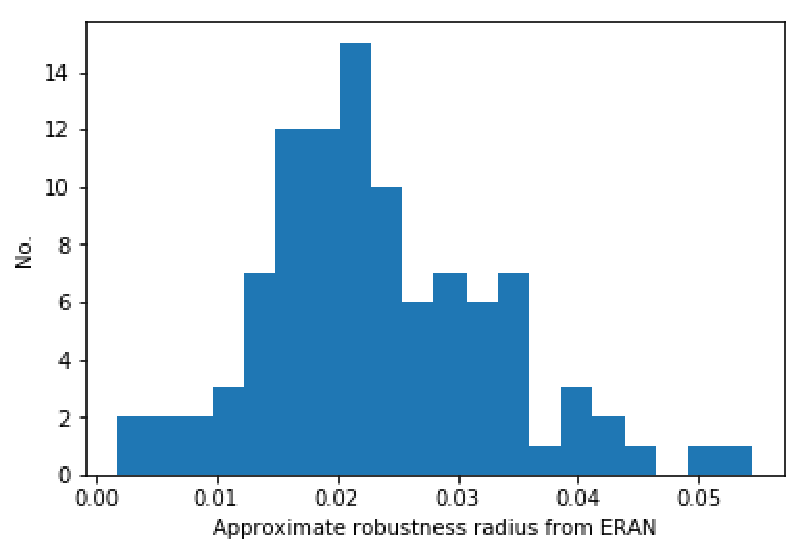}} 
\caption{Distribution of exact/approximate robustness radii of valid inputs}
\label{dis-fnn}
\end{center}
\vskip -0.2in
\end{figure}

Figure~\ref{dis-fnn} shows the histogram of the
density of  exact/approximate robustness
radii from ERAN of the first 100 valid inputs on the network FNN-MNIST.
By observing the figure, the exact and approximate robustness
radii seem follow a similar distribution. One thought is that
they follow a normal distribution. To test that, we  compute the
approximate robustness radii of the first
100 valid inputs for all networks in Table~\ref{p-table}.
We test whether they follow
a normal distribution by D'Agostino and Pearson's test~\cite{d1973tests},
which returns a p-value~\footnote{
In statistics, p-value (or probability value) is
  the probability of obtaining the observed results of a test, assuming
  that the null hypothesis is correct.}
 (shown in column P-value in Table~\ref{p-table}).
  If the p-value
is larger than \(0.05\), then it is possible that they 
follow a normal distribution.  We can see that 25 networks
follow normal distributions,
but 9 not. We cannot give a conclusion on what factors make the difference, but
it seems that a medium size network with high accuracy usually enjoys this property.
It is worth mentioning that the exact robustness radii of the
FNN-MNIST network
do not pass the D'Agostino and Pearson's test either, the
same as its approximate robustness radii, even though they look like a normal distribution in
Figure~\ref{dis-fnn}. Given randomness, not all data in real life can pass
a normality test even if the natural process behind it follows a normal distribution.

\section{Input Validation with Observation II}
If the  approximate robustness radii
of the valid inputs on a network follow a normal distribution, we can
utilize this to validate inputs, which is called
\emph{validation by distribution}.

\begin{algorithm}[tb]
   \caption{Validation by distribution}
   \label{alg:valid}
\begin{algorithmic}
  \STATE {\bfseries Input:} network $\net$, input \(\inputx\)
  \STATE  \(Q = \) a queue (the size of which is \(s\)) of valid inputs
   \REPEAT
   \STATE {\bfseries Input:} input \(\inputx\)
   \STATE \(Q.\mathtt{append}(\inputx)\)
   \IF{\(\algtest(Q)\)}
   \STATE \(\mathtt{del}\) \(Q[0]\)
   \ELSE
   \STATE \(\mathtt{del}\) \(Q[s]\)
    \ENDIF
   \UNTIL{END OF INPUT}
\end{algorithmic}
\end{algorithm}

The new algorithm is shown in Algorithm~\ref{alg:valid}. It maintains a
sliding window~\cite{rebbapragada2009} of size
\(s\) containing the inputs believed to be valid. 
When a new input comes,
the algorithm checks whether it breaks the original normal distribution by
function \(\algtest\). If it does, then this new input is deleted
(\(\mathtt{del}\) \(Q[s]\)), otherwise the first element is deleted (\(\mathtt{del}\) \(Q[0]\)),
thus the window slides.
The design of function \(\algtest\) is heuristic, and we propose the one below

\begin{center}
  \small
  \(
  \begin{array}{l}
    \quad Q[s] \geq \sigma_0  \vee
      \algpvalue(Q[0:s-1]) -  \algpvalue(Q[1:s])  \leq \sigma_1\\
  \end{array}
  \)
\end{center}

This function combines our two observations. It
returns true if the approximate robustness radius of the
last input is larger than a threshold \(\sigma_0\) (from Observation I)
or the p-value of the
new sliding window \(\algpvalue(Q[1:s])\) does not drop sharply from
the last one \(\algpvalue(Q[0:s-1])\) (from Observation II). 
On the  network CNN-MNIST,
we set \(s = 50\), \(\sigma_0 = 0.014\) and \(\sigma_1 = 0.001\)
(the size \(s = 50\) is typical for normality test,
\(\sigma_0\) and \(\sigma_1\) depend on the observation on
the specific neural network).
To test this algorithm, we take 100 valid inputs, 1 misclassified input (because the accuracy
is 98.62\% ), and 100 adversarial examples from each of
the four types of attacks (which totally makes 400) as the inputs.
If all the inputs come in sequence, our algorithm can
reject all adversarial examples and the misclassified input, with only 3 valid
inputs rejected. However, if we shuffle the inputs randomly, the
average rejected valid inputs are 5 and adversarial examples are 28 respectively (by 10 times
experiments). Actually, in both cases,
the first condition (\( Q[s] \geq \sigma_0\)) accepts 87 valid inputs and rejects
all invalid inputs. The second condition accounts for other accepted valid
inputs and false positives.

The experimental results imply that \emph{input validation by distribution} is only suitable for the attack where continuous
adversarial examples are needed.

The disadvantage of this method is that it
needs to compute the approximate robustness radius with Algorithm~\ref{alg:binary}
which needs several iterative calls to \(\algrobust\), which takes more time.
However, the time complexity of  \(\algrobust\) for incomplete verification
is  polynomial wrt. the number of neurons, and the potential
of its speed is far from fully explored (e.g., GPU is not utilized). 





\section{Related Work}


Some researchers focus on finding new adversarial attacks. According to 
whether the attackers have all knowledge about the target neural networks,
adversarial attacks can be divided into two types: white-box attack and black-box
attack. Most  adversarial attacks including the first one
(i.e., L-BFGS~\cite{Szegedy13}) are white-box attacks.
White-box attacks can be fast (e.g., FGSM~\cite{goodfellow2014}) and
strong (i.e., to find the adversarial examples close to the original inputs,
e.g., DeepFool~\cite{moosavi2016}, C\&W~\cite{carlini2017},
Ground-truth attack~\cite{carlini2018}).
Black-box attacks usually need more computational power (e.g., Hopskipjump attack~\cite{chen2019}, ZOO~\cite{chen2017zoo}).
Because of transferability~\cite{papernot2016transferability}, 
white-box attacks can be transformed  to black-box.

There are many countermeasures for adversarial attacks, among which, verification
and adversarial detection are mostly related to our work.

Verification methods check whether a neural network satisfies a given
formally defined property before it is deployed. Such properties include
safety constraints~\cite{katz2017} and robustness~\cite{Ehlers2017,dutta2018}.
However, due to the non-linearity of activation functions, complete
verification is NP-hard, and thus can hardly scale.
Incomplete verification sacrifices the ability to falsify a property
so as to gain performance. Current incomplete verifiers~\cite{wang2018,singh2019}
can deal with neural networks of thousands of neurons in seconds.
However, both verifications can only prove
local robustness properties~\cite{huang2017},
rather than global robustness properties. Thus
these verifiers can only give metrics on evaluating how robust a neural network is, rather
than proving that a neural network is robust.

Adversarial detection methods make use of characterics of adversarial examples.
\cite{feinman2017} found that the uncertainty of adversarial examples to be
higher than clean data, and  utilized a Bayesian neural network to estimate
that. \cite{song2017} found the distribution of adversarial examples is
different from clean data. Compared to their methods focusing on the inputs,
our  method computes the accumulative gradients information of the neural
network in the regions around the inputs. 
\cite{wang2018detect} proposed to detect adversarial examples by mutation test based
on the belief that they are not robust to mutations.
Their method shares the similar intuition with our method, that is,
the adversarial examples must be some corner cases in the input space.
However, we utilize local robustness verification which takes the whole region
around an input into account instead of testing which considers some points
near an input.
\cite{lu2017} distinguishes adversarial examples
from clean data by the threshold of their values on each ReLU layer.
\cite{henzinger2019} proposed to detect novel inputs by observing the hidden
layers, i.e., whether they are outside the value ranges during training.

\emph{Comparison with other works on effectiveness.}
Given the fact that these aforementioned works are not open source and the results in their papers are often
given in the form of graphs (like ROC curve), it is hard to have a fair comparison with
their results. However, from the results in their papers, our method is comparable (if not
better) with their work, especially on strong attacks.

\section{Conclusion}

Neural networks are known to be vulnerable to adversarial examples,
which can cause serious safety and security issues. Current
techniques can hardly verify the robustness of a neural network
for all possible inputs. Thus to detect adversarial examples
at run-time is an important way to protecting the systems with neural networks.

We believe that a well trained neural network does not
leave big "holes" for natural misclassified inputs
and adversarial examples, which results in small robustness radii
of these inputs.
Based on this belief, we conducted empirical
study on the exact/approximate robustness radii of valid inputs,
natural misclassified  inputs and adversarial examples from
various attacks. We have the  following two observations:
(1) the exact/approximate robustness radii of valid inputs
are much larger than those of misclassified inputs and adversarial
examples, especially those from strong attacks;
(2) the approximate robustness radii of
valid inputs often follow a normal distribution.

Based on the two observations, we propose to
leverage local robustness verification to
validate inputs. The first method is validation by threshold,  which can  improve the accuracy
of neural networks and protecting them from adversarial examples. Moreover,
this method allows the users to set the threshold as they wish based on
the maximal false alarm rate that can be tolerated. The second method is
validation by distribution. It is suitable for detecting continuous attacks.
Our experiments show that both methods are effective.

\section{Acknowledgement}
This work has been partially supported by  National Natural Science Foundation of China (Grant No.62002363), and the Natural Science Foundation of Hunan Province of China (Grant No. 2021JJ40698).

\bibliographystyle{plain}

\bibliography{paper}

\end{document}